\documentclass[10pt,twocolumn,letterpaper]{article}

\usepackage{cvpr}              % To produce the CAMERA-READY version
\definecolor{cvprblue}{rgb}{0.21,0.49,0.74}
\usepackage[pagebackref,breaklinks,colorlinks,allcolors=cvprblue]{hyperref}
\usepackage{booktabs}
\usepackage{array}
\usepackage{soul} % 형광펜 효과를 위한 패키지
\usepackage{xcolor} % 색상 추가
\usepackage{graphicx}

\usepackage{pifont}   
\usepackage{xcolor}   
\usepackage{multirow}

\usepackage[table]{xcolor}
\usepackage{makecell}
\usepackage{arydshln}
\usepackage{tcolorbox}

\def\paperID{-} % *** Enter the Paper ID here
\def\confName{CVPR}
\def\confYear{2026}

\title{Can Language Models Understand mmWave Data? Benchmarking Large Language Models for mmWave Radar-Based Human Understanding}

\author{
Jeongwan Shin$^{1,2}$ \quad
Jaehyeon Kim$^{1}$ \quad
Donguk Ko$^{1}$ \quad
Jaeho Choi$^{\dag,1,2}$ \quad \\
$^{1}$DGIST \quad 
$^{2}$KAIST InnoCORE LLM \\
}

\begin{document}
\twocolumn[{%
\renewcommand\twocolumn[1][]{#1}%
\maketitle
\begin{center}
    \centering
    \captionsetup{type=figure}
    \includegraphics[width=\linewidth, keepaspectratio]{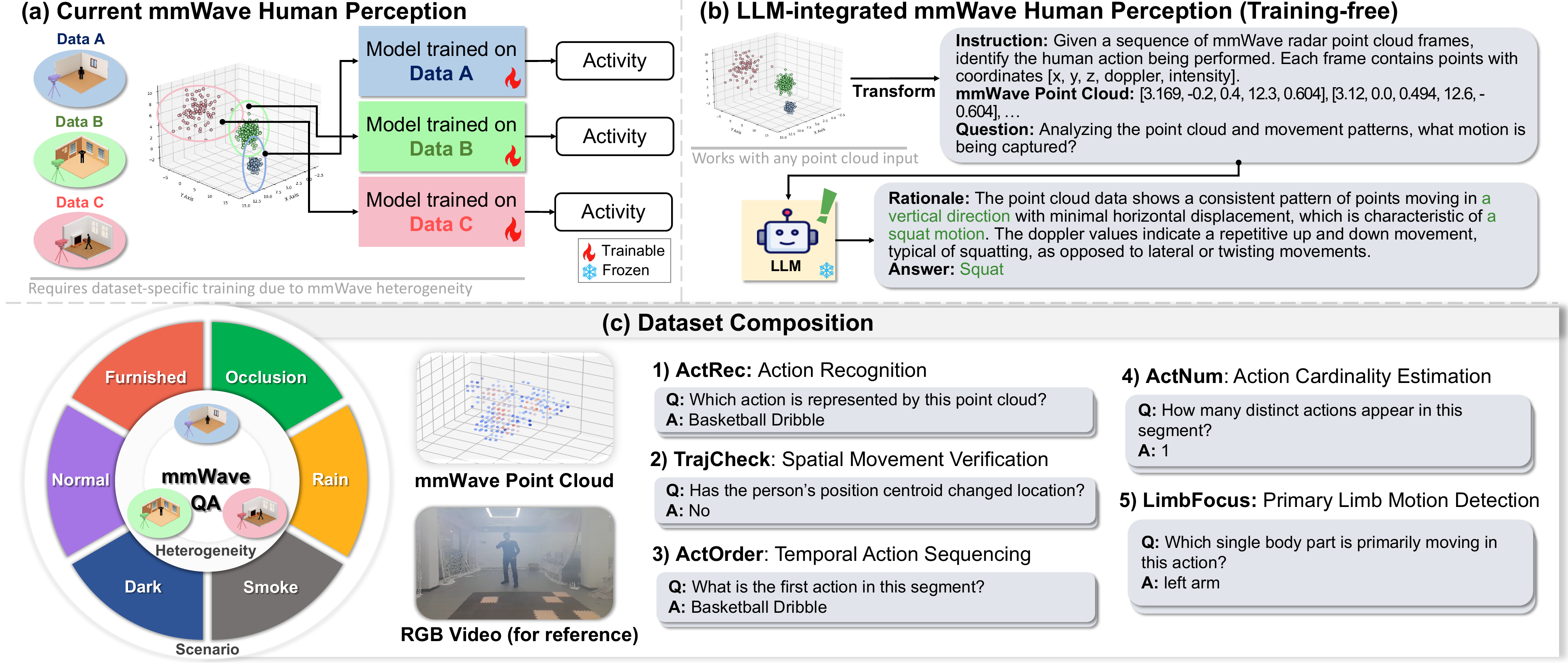} 
    \caption {\textbf{Top:} Conventional mmWave perception frameworks relying on per-dataset retraining vs. our LLM-based perception that interprets structured radar point cloud inputs in a zero-shot manner.
    \textbf{Bottom:} Composition of mmWave-QA, encompassing heterogeneous mmWave devices, six real-world scenarios, and five QA types for comprehensive multimodal reasoning assessment.}
  \label{fig:fig1} 
\end{center}%
}]
\maketitle
\begin{abstract}
Large language models (LLMs) have shown remarkable reasoning and generative capabilities, motivating their use as universal reasoning engines for perception. 
While modern approaches such as vision-language models (VLMs) have attempted to incorporate reasoning capabilities into visual sensing, the integration of LLMs with the millimeter-wave (mmWave) modality--despite its unique advantages under low light and occlusion--remains largely unexplored. 
The principal bottlenecks stem from the scarcity of radar-language pairs, severe cross-dataset heterogeneity, and the absence of a foundational mmWave encoder. 
We address this gap through a minimal textualization interface that serializes each mmWave point cloud into concise natural language, allowing off-the-shelf LLMs to operate in a question-answering (QA) setting.
Building on this, we present mmWave-QA, the first benchmark for language-conditioned mmWave human perception. 
mmWave-QA aggregates heterogeneous public mmWave datasets and harmonizes them via calibration-aware preprocessing and global taxonomy alignment, while providing natural language QA.
Spanning six scenarios and five QA tasks, the benchmark enables standardized evaluation across diverse mmWave hardware and experimental conditions, establishing a foundation for scalable research on mmWave-LLM integration. 
We further evaluate and analyze LLMs on our mmWave-QA, highlighting their zero-shot reasoning potential for radar perception, as well as their robustness under visual degradation.

\begingroup
\renewcommand\thefootnote{}\footnotetext{
    \textsuperscript{\dag}Corresponding author.}
\endgroup

\end{abstract}    
\vspace{-5pt}
\section{Introduction}
\label{sec:intro}

Millimeter-wave (mmWave) sensing has emerged as a strong alternative for human perception due to its ability to measure human targets via electromagnetic reflections rather than appearance cues \citep{mmholmes, mri, HuPR, mID}.
By capturing human kinematics across range, Doppler, and angle domains, modern mmWave radar systems remain effective even under challenging conditions that routinely degrade cameras--low/mixed lighting, partial/strong occlusions of the human body, and long standoff distances \citep{MVDoppler-pose, mmMesh}. 
Furthermore, the mmWave signals do not reveal facial identity or fine appearance, naturally offering a privacy advantage that is essential for continuous and ubiquitous monitoring in everyday home or office environments \citep{ProbRadarM3F, MU-ID, Alloulah_2023_CVPR}. 

Meanwhile, recent LLMs \citep{llama, mistral, GPT3, openai2023gpt4, openai2024gpt4o} trained on hyper-scale corpora \cite{BigScience, RefinedWeb, Dolma} have demonstrated powerful contextual understanding \citep{zhu-etal-2024-large, zhao-etal-2024-enhancing, ravaut-etal-2024-context} and broad generalization across diverse tasks \citep{multitask, feedback, Scaling, VisualIstTuning, Liu_2024_CVPR, zeroVQA, videoLlama, llavaVideo}. 
When further integrated with visual encoders \citep{BLIP, clip, siglip}, these models evolve into VLMs \citep{Flamingo, pali, Internvl}, where perception is conditioned on both linguistic and visual cues, allowing language-guided scene understanding with minimal supervision \citep{ovseg, GroupViT}. 
This context-conditioned paradigm offers two notable benefits for human perception: 1) a natural-language interface that enables humans to query and steer models through QA-based interactions and 2) the ability to leverage extensive knowledge of LLMs to interpret unseen behaviors in a zero-/few-shot manner \citep{GPT3, CoT, self-cons}.

However, integrating LLMs with the mmWave modality presents several non-trivial challenges.
First, an mmWave–language ecosystem remains underdeveloped: unlike common modalities (\eg, RGB or audio) with web-scale modality-language pairings, existing mmWave datasets are small, lab-specific, and hard to annotate, leading to the absence of public mmWave datasets with radar-caption pairings. 
Second, this data scarcity is further compounded by heterogeneity: mmWave observations vary not only due to environmental clutter/multipath but also by hardware-specific factors (\eg, carrier frequency, waveform design, $\#$ antennas), which in turn induce severe distributional shifts across datasets \citep{RF, mtranssee, mmGPE, FastScalableHPE}. 
Third, the field lacks standardized tokenizers or foundational encoders to bridge mmWave tensors and language backbones.
Consequently, current mmWave systems for human understanding \citep{MU-ID, RadarAE, RF-pose, RF-Pose3D, M4esh} remain scenario-tuned: models are optimized for each dataset and require retraining whenever the scenario (\eg, hardware, surroundings, or subject distribution) changes, as examplified in \cref{fig:fig1} (a). 
Accordingly, our work begins with a concrete question: \textit{can we still leverage the reasoning power of LLMs for mmWave-based human understanding despite the aforementioned challenges?}      

To investigate its feasibility, we render each 5D mmWave point cloud sample into a coordinate-formatted text representation (\ie, [$x$, $y$, $z$, Doppler, intensity]) and prompt LLMs to answer the corresponding human motion question (\cref{fig:fig1} (b)). 
Surprisingly, without any mmWave-specific tuning, LLMs exhibit reasonable zero-shot reasoning, reflecting their potential to map textualized radar data to high-level behaviors. 
Building on this observation, we introduce mmWave-QA, the first benchmark designed to explore mmWave-LLM integration. 
The benchmark is constructed through a curation pipeline that integrates heterogeneous datasets, reformulates them into mmWave-based QA tasks aligned with the benchmark taxonomy, and generates QA pairs through an LLM–human collaboration process.
Finally, quality review and refinement are conducted to ensure the quality and balance of the resulting benchmark.
After constructing the benchmark, we evaluate various prompting strategies to comprehensively assess and analyze how LLMs interpret human motion from mmWave data.

As shown in \cref{fig:fig1} (c), our mmWave-QA provides a unified testbed spanning six diverse scenarios with five complementary QA tasks that go beyond simple spatial action queries to probe spatial, temporal, and joint-level granular aspects of human motion.
Furthermore, our benchmark encompasses different mmWave hardware devices and experimental conditions, thereby enabling standardized evaluation in analyzing mmWave-LLM integration under diverse co-factors:  subject placements, environment/scenario shifts, and even hardware/waveform dependencies.
Through extensive experiments on mmWave-QA, we investigate the behavior of off-the-shelf LLMs \citep{openai2024gpt4o, deepmind2025gemini25} under different prompting strategies, revealing their ability to interpret mmWave-based human motion without task-specific fine-tuning.
We further evaluate the potential of mmWave–LLM integration for zero-shot human understanding, demonstrating that the mmWave modality remains more robust than RGB under visual degradation.

In summary, our contributions are threefold:
\begin{itemize}
    \item We present mmWave-QA, a unified benchmark that harmonizes heterogeneous mmWave human-motion datasets and recasts them as natural-language QA tasks spanning multiple aspects of action understanding.
    \item We show that integration with simple off-the-shelf LLMs can even lead to strong zero-shot reasoning on textualized mmWave radar inputs.
    \item We provide empirical evidence that the mmWave modality is more robust than RGB under visual degradation.
\end{itemize}

\begin{table*}[t]
\centering
\footnotesize
\setlength{\tabcolsep}{2pt}
\renewcommand{\arraystretch}{0.6}
\begin{tabular}{>{\raggedright\arraybackslash}p{2cm}>{\centering\arraybackslash}p{0.6cm}>{\centering\arraybackslash}p{0.9cm}>{\centering\arraybackslash}p{0.6cm}>{\centering\arraybackslash}p{1.2cm}>{\centering\arraybackslash}p{1.2cm}>{\centering\arraybackslash}p{1.1cm}>{\centering\arraybackslash}p{0.7cm}>{\centering\arraybackslash}p{0.7cm}>{\centering\arraybackslash}p{0.7cm}>{\centering\arraybackslash}p{0.7cm}>{\centering\arraybackslash}p{0.7cm}>{\centering\arraybackslash}p{0.7cm}>{\centering\arraybackslash}p{0.4cm}>{\centering\arraybackslash}p{0.4cm}>{\centering\arraybackslash}p{0.4cm}>{\centering\arraybackslash}p{0.5cm}>{\centering\arraybackslash}p{0.5cm}>{\centering\arraybackslash}p{0.5cm}}
\toprule
\multirow{2}{*}{\textbf{Benchmarks}}    & \multirow{2}{*}{\textbf{\#Act}} & \multirow{2}{*}{\textbf{\#Subj}} & \multirow{2}{*}{\textbf{\#Env}} & \multirow{2}{*}{\textbf{mmWave}} & \multirow{2}{*}{\textbf{Sensor}} & \multirow{2}{*}{\textbf{Dataset}} & \multicolumn{6}{c}{\textbf{Scenario}} & \multicolumn{3}{c}{\textbf{Modality}} & \multicolumn{3}{c}{\textbf{Annotations}} \\  \cmidrule(lr){8-13} \cmidrule(lr){14-16} \cmidrule(lr){17-19} 
& & & & \textbf{HW} & \textbf{Res.} & \textbf{Size} & Norm. & Furn. & Rain & Smoke & Dark & Occl.  & R & V & T & Act & Pose & QA 
\\ 
\midrule
RadHAR~\citep{RadHAR} & 5 & 2 & 1 & A & Low & M &\checkmark & - & - & - & - & - & \checkmark & - & - & \checkmark & - & - \\
MVDoppler~\citep{MVDoppler} & 4 & 13 & 1 & B & Low & - &\checkmark & - & - & - & - & - & \checkmark & - & - & \checkmark & - & - \\
XRF55~\citep{XRF55} & 55 & 39 & 4 & C & Low & XL &\checkmark & - & - & - & - & - & \checkmark & \checkmark & - & \checkmark & - & - \\
MARS~\citep{MARS} & 10 & 4 & 1 & A & Low & S &\checkmark & - & - & - & - & - & \checkmark & - & - & - & \checkmark & - \\
mmBody~\citep{mmbody} & \textit{100} & 20 & 6 & D & High & M &\checkmark & \checkmark & \checkmark & \checkmark & \checkmark & \checkmark & \checkmark & \checkmark & - & - & \checkmark & - \\
HIBER~\citep{HIBER} & 4 & 10 & \textit{10} & E & High & M &\checkmark & - & - & - & - & \checkmark & \checkmark & \checkmark & - & - & \checkmark & - \\
MMVR~\citep{MMVR} & - & 25 & 6 & E & High & M &\checkmark & - & - & - & - & \checkmark & \checkmark & \checkmark & - & - & \checkmark & - \\
mRI~\citep{mri} & 12 & 20 & 1 & A & Low & M &\checkmark & - & - & - & - & - & \checkmark & \checkmark & - & \checkmark & \checkmark & - \\
MM-Fi~\citep{mmFi} & 27 & \textit{40} & 4 & C & Low & M & \checkmark & - & - & - & - & - & \checkmark & \checkmark & - & \checkmark & \checkmark & - \\ 
\midrule
\textbf{Ours} & \textbf{139(86)} & \textbf{80} & \textbf{11} & \begin{tabular}[c]{@{}c@{}}A, C, D \end{tabular} & Low, High & L & \checkmark & \checkmark & \checkmark & \checkmark & \checkmark & \checkmark & \checkmark & \checkmark & \checkmark & \checkmark & \checkmark & \checkmark \\ 
\bottomrule
\end{tabular}
%\caption{Comparison of mmWave-based human sensing benchmarks (A:\textcolor{red}{(TI)} IWR1443, B: AWR1843, C: IWR6843, D: Phoenix type, E: AWR2243). Numbers in parentheses denote the number of newly categorized high-level action classes.}
\caption{
\textbf{Comparison of mmWave-based human sensing benchmarks.} 
Each column denotes the following -- 
number of distinct human actions (\#Act), 
number of subjects (\#Subj), 
number of environmental settings (\#Env), and 
radar hardware type (mmWave HW):
A) TI IWR1443, B) TI AWR1843, C) TI IWR6843, D) Phoenix-type custom board, E) TI AWR2243. 
Scenario cover diverse sensing conditions including normal, furnished, rain, smoke, dark and occlusion environments. 
Modality specifies the available input types: radar, vision, and text. 
Annotations indicate the provided supervision signals: 
human action labels (Act), 
3D pose annotations (Pose), and 
natural-language QA pairs (QA). 
Numbers in parentheses (\eg, (86)) represent newly categorized action classes.
Datasets are categorized by the number of frames: S ($<$100k), M (100–400k), L (400–700k), and XL ($>$700k), as detailed in the supplementary material.
}
\label{tab:table1}
\end{table*}

\section{Related Work}
\label{sec:relatedwork}
\paragraph{mmWave Human Sensing Datasets.}
As shown in Table~\ref{tab:table1}, several mmWave human sensing datasets have been introduced in recent years. 
Within action recognition, datasets have evolved from RadHAR \citep{RadHAR} and MVDoppler \citep{MVDoppler}, which capture basic body motions to XRF55 \citep{XRF55}, which covers composite actions and diverse environments.
Extending beyond action recognition, pose estimation \citep{MARS, mri, mmFi} focuses on joint-level understanding, spanning from rehabilitation to natural daily motions, and further enriches datasets by expanding the number of subjects and environments.
With recent advances in radar hardware, mmBody~\citep{mmbody}, HIBER~\citep{HIBER}, and MMVR~\citep{MMVR} employ high-resolution radar sensors, providing denser and more detailed observations.
Despite these advances, the growing heterogeneity in data sources and sensing setups continues to hinder the development of universally adaptable models.

\paragraph{mmWave-based Human Understanding.}
Considering the diversity of human sensing data, mmWave-based human understanding has evolved into several distinct modeling paradigms.
MARS \citep{MARS} and mmPose \citep{mmPose} employ CNN-based architectures that learn spatial representations from radar reflections, while RadHAR \citep{RadHAR} and DP-CBL \citep{DP-CBL} further incorporate LSTM to capture temporal dependencies across consecutive frames. 
Inspired by the success of transformers in achieving powerful context understanding across NLP \citep{BERT} and vision \citep{ViT, SwinViT, mobileViT} domains, HMR-SCL \citep{HMR-SCL}, MVDoppler-Pose \citep{MVDoppler-pose}, and mmPoint \citep{mmPoint} adopt attention mechanisms, underscoring how mmWave encoding paradigms largely track broader architectural trends in other domains.
Despite growing applications of LLMs for contextual reasoning, their integration with mmWave sensing remains limited; mmWave-QA addresses this gap through a benchmark for systematic exploration. 

\paragraph{Sensing modalities with LLMs.}
By incorporating additional modalities, LLMs can be extended into multimodal, allowing them to interpret multimodal inputs such as images \citep{Liu_2024_CVPR, InstructBLIP, VisualIstTuning, BLIP}, videos \citep{llavaVideo, VIMI, videoLlama}, audio \citep{Lu_2024_CVPR, Geng_2025_CVPR}, and spatial measurements \citep{PointLLM, 11086426}.
Building on this progress, only a limited number of studies have explored mmWave sensing modalities.
Among recent attempts, RadarLLM \citep{radarLLM} is the framework that leverages LLMs for human motion understanding using millimeter-wave radar, training on synthetically generated IF signals to bridge radar sensing and language understanding.
Similarly, HoloLLM \citep{holoLLM} introduces a multisensory foundation model that integrates sensing modalities with LLMs for language-grounded human perception and reasoning.
However, existing models face inherent limitations: those relying on synthetic radar data struggle to generalize to real-world clutter, while those processing raw mmWave signals depend on modality-specific encoders.
To overcome these limitations, we demonstrate that existing LLMs can interpret textualized mmWave point clouds in a zero-shot manner and establish a benchmark that quantifies their reasoning capability, laying the groundwork for future radar–language integration.
\begin{figure*}[t] 
  \centering
  \includegraphics[width=\linewidth]{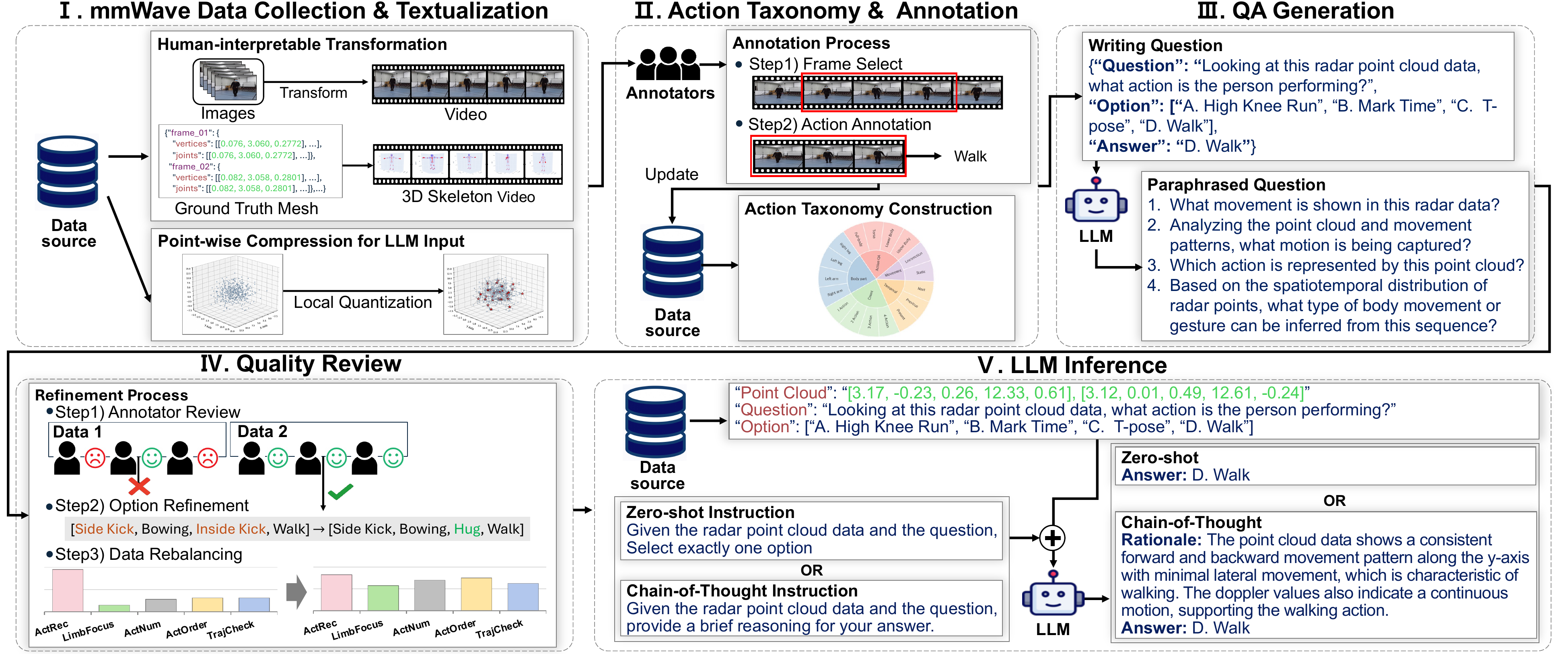} 
    \caption{
    \textbf{Pipeline of mmWave-QA construction and evaluation:}
    I) Original mmWave Collection \& Textualization--radar data are preprocessed into human-interpretable and LLM-ready quantized formats;
    II) Action Taxonomy \& Annotation--actions are categorized into hierarchical domains; 
    III) QA Generation--diverse QA pairs are generated through human-in-the-loop LLM; 
    IV) Quality Review--annotators refine and balance QA; 
    and V) LLM Inference--the benchmark enables evaluation of language models across diverse prompting.
    }
  \label{fig:pipeline}
\end{figure*}

\section{mmWave-QA Benchmark}
\label{sec:mmWaveQA}

\subsection{Data Construction}

As a step toward building the first mmWave–language benchmark, we integrate three mmWave sensing datasets for human perception: mmBody \citep{mmbody}, MM-Fi \citep{mmFi}, and mRI \citep{mri}.
This integration provides a comprehensive and diverse foundation covering multiple radar devices, environments, and subjects.
Merging the datasets yielded 139 action labels, which were refined into 86 distinct categories after removing redundancies and ambiguities.Table~\ref{tab:table1} compares mmWave-QA with prior human sensing benchmarks, highlighting its diversity and QA-driven design, while \cref{fig:pipeline} illustrates the overall dataset construction pipeline.

\paragraph{mmWave Data Collection and Textualization.}
We begin by addressing the insufficiencies of prior mmWave datasets, which contained dense and cluttered point clouds with limited visual information.
RGB recordings are converted into visual references for annotators, presenting clean scenes as videos and degraded or occluded ones as skeleton videos to aid human understanding.
To standardize data representation across heterogeneous point-cloud formats, we apply coordinate-based quantization to cluster points into a form compatible with LLM token limits.

\begin{figure*}[t] 
  \centering
  \includegraphics[width=\linewidth]{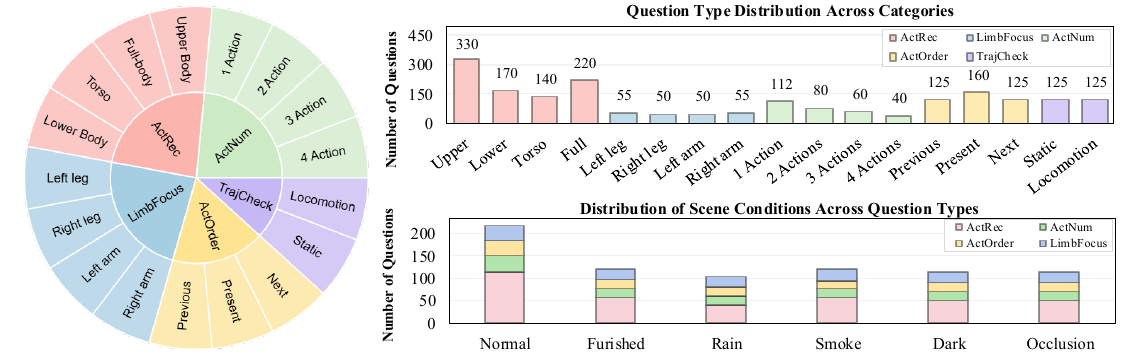} 
  \caption{
        \textbf{Statistics of mmWave-QA dataset.} 
        \textbf{Left:} Hierarchical taxonomy of question domains: ActRec, TrajCheck, ActOrder, ActNum, and LimbFocus. 
        \textbf{Top-right:} Distribution of subcategories within each top-level action domain, illustrating diverse coverage across human-motion questions. 
        \textbf{Bottom-right:} Distribution of taxonomy actions across six scene conditions: normal, furnished, rain, smoke, dark, and occlusion.
        Note that the TrajCheck category is omitted in the bottom-right plot due to limited samples under specific conditions.
    }
  \label{fig:statistics}
\end{figure*}

\vspace{-10pt}
\paragraph{Action Taxonomy and Annotation.}
As some of the original datasets do not provide action-wise segmentation, annotators segment the videos into continuous action intervals, assigning the corresponding action label.
Based on the annotated action segments, annotators construct a hierarchical taxonomy of question types, as illustrated on the left side of \cref{fig:statistics}. 
Specifically, mmWave-QA comprises five core categories that capture different aspects of human motion understanding, enabling multifaceted evaluation on how well an mmWave-conditioned LLM can perceive and align physical information with linguistic representations.
%\textcolor{red}{Specifically, mmWave-QA comprises five core categories that capture different aspects of human motion understanding, which is essential for enabling human-centric LLMs to perceive and respond to physical behaviors through mmWave sensing}:
(I) Action Recognition (ActRec) for recognizing the performed action,
(II) Spatial Movement Verification (TrajCheck) for determining whether the subject’s centroid has changed position,
(III) Temporal Action Sequencing (ActOrder) for reasoning about the temporal relationships between consecutive actions,
(IV) Action Cardinality Estimation (ActNum) for estimating the number of distinct actions, and
(V) Limb Motion Detection (LimbFocus) for identifying the primary moving body part.
Each category is divided into subtypes capturing motion attributes.
%Each category is further divided into fine-grained subtypes, capturing specific motion attributes such as body region, temporal order, or action count.

%The distributions of these QA categories across domains are shown in the upper-right panel of \cref{fig:statistics},
%while the lower-right plot illustrates their distribution under different scene conditions, including public, furnished, rain, smoke, poor lighting, and occlusion environments.
%Furthermore, our benchmark includes paired RGB videos alongside mmWave data to enable cross-modal comparisons under diverse real-world conditions. 
%To comprehensively assess robustness, we construct QA pairs across various scene settings such as \textit{public}, \textit{occlusion}, \textit{poor lighting}, \textit{furnished}, \textit{rain}, and \textit{smoke}. 
%As illustrated in the lower-right of Figure~\ref{fig:2}, visual data often exhibit degraded reliability under these conditions due to visibility or privacy constraints, whereas mmWave sensing remains robust and privacy-preserving. 
%This design highlights the complementary nature of mmWave-based reasoning beyond the limitations of conventional visual understanding.

\paragraph{Question–Answer Generation.}
Following the annotation stage, annotators generate natural language question–answer pairs based on the labeled actions.
%Building on the annotated actions, annotators generate natural language QA pairs.
%\textcolor{red}{As no prior work has provided natural-language QA pairs for mmWave data, we build upon the annotated actions to generate such pairs.}
Each question is designed to evaluate different aspects of point cloud understanding in accordance with the taxonomy established in the previous step.
Subsequently, annotators construct multiple-choice QA pairs by selecting answer options from the predefined action taxonomy.
To enhance linguistic diversity and ensure natural phrasing, we further employ LLMs leveraging its paraphrastic generation capability \citep{openai2023gpt4, luo2023vectorquantized} to perform paraphrasing of the original questions while preserving their semantic intent.
%The resulting QA pairs thus cover a broad range of reasoning skills and linguistic expressions, forming the foundation of our mmWave-QA Benchmark.

\vspace{-10pt}
\paragraph{Quality Review.}
Existing human motion datasets often suffer from biased action distributions and inconsistent labeling of visually similar actions. To ensure annotation reliability and dataset balance, we conduct a three-step quality review process.
%\textcolor{red}{Human motion datasets often exhibit biased action categories and inconsistent annotations across similar actions.}
%To guarantee the reliability and balance of annotations, we perform a three-step quality review process. 
Three annotators independently review all generated QA pairs, and any pair that appears ambiguous or contains unclear action semantics is removed from the dataset if even one annotator marks it as invalid. 
During this step, we refine the answer options by replacing overly similar or redundant candidates with more distinctive ones to prevent confusion during QA.
Finally, we adjust the distribution of QA pairs to ensure a balanced coverage across different action domains and question types defined in the taxonomy.
After completing refinement procedures, we finalize the mmWave-QA benchmark and perform evaluation using LLM inference under diverse prompting strategies.
%Finally, because the number of samples varies across categories in the predefined taxonomy, we adjust the sampling ratios to achieve a balanced distribution across different action domains and question types.

\subsection{Dataset Statistics}
\label{sec:statistics}
The top-right chart in \cref{fig:statistics} presents the distribution of question types across domains in mmWave-QA.
Among all categories, ActRec and TrajCheck questions occur most frequently, whereas ActOrder, ActNum, and LimbFocus appear less often yet enrich the benchmark by addressing sequential reasoning, quantitative estimation, and localized motion analysis.
Together, these categories provide a comprehensive, balanced, and complementary set of tasks for evaluating multimodal reasoning capabilities of language models.
The bottom-right chart presents the distribution of question types across six scene conditions: normal, furnished, rain, smoke, dark, and occlusion.
Note that the TrajCheck category is excluded due to an insufficient number of samples in certain conditions.
Each condition contains roughly 100–200 questions, with Normal scenes having the largest share due to their higher data availability and role as the baseline for comparison.
This diversity enables consistent evaluation of model robustness and generalization across heterogeneous environmental settings.

\section{LLM-driven mmWave-QA}
\label{sec:LLM_Inference}
\subsection{Preliminary}
\label{sec:preliminary}
We first formalize the inference process of an LLM on the proposed mmWave-QA benchmark. 
Given a natural-language question $Q$ and a mmWave point cloud sequence $P = \{p_i\}_{i=1}^{l}$ consisting of $l$ frames, 
where each frame $p_i$ exhibits varying point density depending on the sensing device and environmental conditions,
we formally define the inference process of an LLM $\mathcal{M}$ as:
\begin{equation}
    A = \mathcal{M}(I, Q, P),
\end{equation}
where $I$ denotes the instruction prompt defining the inference setting, and $A$ represents the sequence of output tokens generated by $\mathcal{M}$. 
This formulation serves as the foundation for subsequent prompting strategies, including zero-shot, few-shot, and chain-of-thought inference.

\begin{table*}[t]
\centering
\small
\renewcommand{\arraystretch}{0.5}
\setlength{\tabcolsep}{3pt}
\begin{tabular}{>{\raggedright\arraybackslash}p{2.2cm}>{\raggedright\arraybackslash}p{2cm}>{\centering\arraybackslash}p{1.6cm}>{\centering\arraybackslash}p{1.6cm}>{\centering\arraybackslash}p{1.6cm}>{\centering\arraybackslash}p{1.6cm}>{\centering\arraybackslash}p{1.2cm}>{\centering\arraybackslash}p{1.2cm}>{\centering\arraybackslash}p{1.2cm}>{\centering\arraybackslash}p{1cm}}
\toprule
\multirow{2}{*}{\textbf{Model}} & \multirow{2}{*}{\textbf{Prompt Type}} &
\multicolumn{4}{c}{\textbf{Action Category}} &
\multicolumn{3}{c}{\textbf{mmWave HW}} & \multirow{2}{*}{\textbf{Total}}\\
\cmidrule(lr){3-6} \cmidrule(lr){7-9}
& &
\makecell[c]{Upper-body} &
\makecell[c]{Lower-body} &
\makecell[c]{Torso} &
\makecell[c]{Full-body} &
Phoenix & IWR6843 & IWR1443 & \\
\midrule
\multirow{4}{*}{\textbf{Gemini2.5-flash}}
& Zero-shot     & 32.12          & 23.53          & 25.00          & 29.09           & 28.06          & 29.14          & 30.77          & 28.49\\
& Few-shot      & 45.76          & 24.12          & 25.71          & 29.55           & 34.03          & 34.29          & 33.85          & 34.07\\
& CoT           & 37.27          & 25.29          & 28.57          & 30.91           & 32.10          & 32.00          & 29.23          & 31.86\\
& Few-shot CoT  & 46.97          & \textit{30.59} & 27.14          & 29.55           & 36.13          & 36.00          & 35.38          & 36.05\\
\midrule
\multirow{4}{*}{\textbf{Gemini2.5-Pro}}
& Zero-shot     & 32.73          & 24.71          & 25.71          & 28.18           & 27.90          & 29.71          & 35.38          & 28.84\\
& Few-shot      & 46.36          & 26.47          & 27.14          & 29.55           & 35.16          & 33.71          & 36.92          & 35.00\\
& CoT           & 41.52          & 24.71          & 27.86          & 29.09           & 32.90          & 31.43          & 35.38          & 32.79\\
& Few-shot CoT  & \textit{47.88} & 28.24          & 28.57          & 30.91           & 36.45          & \textit{36.57} & \textit{36.92} & 36.51\\
\midrule
\multirow{4}{*}{\textbf{GPT-4o}}
& Zero-shot     & 29.09          & 20.59          & 24.29          & 28.18           & 26.13          & 27.43          & 26.15          & 26.40\\
& Few-shot      & 45.45          & 26.47          & 27.86          & 30.91           & 36.77          & 31.43          & 29.23          & 35.12\\
& CoT           & 34.85          & 22.35          & 25.00          & 29.09           & 29.19          & 30.29          & 27.69          & 29.30\\
& Few-shot CoT  & 46.97          & 28.82          & 29.29          & \textit{33.18}  & \textit{38.39} & 33.71          & 32.31          & \textit{36.98}\\
\midrule
\multirow{4}{*}{\textbf{GPT-5}}
& Zero-shot     & 32.12          & 21.76          & 23.57          & 28.18           & 26.45          & 27.43          & 26.15          & 26.63\\
& Few-shot      & 45.76          & 28.82          & \textit{30.00} & 31.36           & 37.42          & 33.71          & 30.77          & 36.16\\
& CoT           & 35.45          & 23.53          & 27.14          & 29.55           & 29.84          & 32.00          & 29.23          & 30.23\\
& Few-shot CoT  & \textbf{48.48} & \textbf{31.76} & \textbf{31.43} & \textbf{33.64}  & \textbf{38.71} & \textbf{37.14} & \textbf{38.46} & \textbf{38.37}\\
\bottomrule
\end{tabular}
\caption{Evaluation of various LLMs across different prompt settings and action categories on multiple mmWave HW.  
    All values denote accuracy (\%), with the best results in \textbf{bold} and the second-best in \textit{italic}
    }
\label{tab:action_category_comparison}
\end{table*}

\subsection{Inference Strategies}
\label{sec:prompting}
We investigate three prompting inference strategies to evaluate the reasoning performance of LLMs on the proposed mmWave-QA benchmark: zero-shot, few-shot, and chain-of-thought (CoT) inference. 
Each strategy conditions the model with different inputs, leading to distinct reasoning processes and answer generation behaviors.

\paragraph{Zero-shot.}
In the zero-shot setting, the model generates an answer directly from the given question–point cloud pair $(Q, P)$ without additional examples or reasoning context:
\begin{equation}
    A_{\text{zero}} = \mathcal{M}_{\text{zero}}(I, Q, P),
\end{equation}
where $A_{\text{zero}}$ denotes the generated answer, 
and $\mathcal{M}_{\text{zero}}$ represents the LLM conditioned on zero-shot prompting.

\paragraph{Few-shot.}
%Given few-shot in-context examples consisting of question–point cloud–answer triplets ${(Q_i, P_i, A_i)}_{i=1}^{k}$ as references for inference:
%Under the few-shot prompting setting, the model is provided with $k$ in-context examples consisting of question–point cloud–answer triplets ${(Q_i, P_i, A_i)}_{i=1}^{k}$ as references for inference:
%Given few-shot in-context examples, the model is provided with $k$ in-context examples consisting of question–point cloud–answer triplets $\{(Q_i, P_i, A_i)\}_{i=1}^{k}$ as references for inference:
Given $k$ few-shot in-context examples consisting of question–point cloud–answer triplets ${(Q_i, P_i, A_i)}_{i=1}^{k}$ as contextual references for inference:
\begin{equation}
    A_{\text{few}} = \mathcal{M}_{\text{few}}(I, \{(Q_i, P_i, A_i)\}_{i=1}^{k}, Q, P),
\end{equation}
enabling the LLM to utilize these examples to infer answers for the target input without updating model parameters.

\paragraph{Chain-of-Thought (CoT).}
Under the CoT prompting setting, the model follows a rationale-generation instruction to produce a continuous text sequence that first presents intermediate reasoning steps before providing the final answer.
%In the CoT setting, the model is guided by a rationale-generation instruction to produce a continuous text sequence that first articulates an intermediate rationale and then provides the final answer.
Formally, the CoT inference can be expressed as:
\begin{equation}
    R_{\text{CoT}} = \mathcal{M}_{\text{CoT}}(I_{\text{rationale}}, Q, P),
\end{equation}
where $I_{\text{rationale}}$ represents the rationale-generation instruction, and $R_{\text{CoT}}$ denotes the LLM-generated sequence $[\text{rationale}; A_{\text{CoT}}]$.
The CoT process enables the model to generate a rationale that explains inter-frame relationships within point clouds, thereby facilitating the derivation of the final answer.
Lastly, we additionally introduce a hybrid inference strategy that integrates few-shot examples with CoT reasoning.
All prompt examples and detailed cases are presented in the supplementary material.

\section{Experiments}
\label{sec:experiments}
%The proposed mmWave-QA benchmark is employed to assess the reasoning capabilities of LLMs on radar-based human activity data collected under heterogeneous devices, environments, and scene conditions.
%The proposed mmWave-QA benchmark evaluates LLM reasoning on radar-based human activity data collected under heterogeneous devices, environments, and scene conditions.
%Evaluation settings are described first, followed by quantitative results under different prompting strategies, along with analyses comparing mmWave- and RGB-based QA and examining the effect of frame count on performance.
%Finally, qualitative examples are presented to illustrate how LLMs reason over mmWave point cloud inputs under diverse scene conditions. 
%Finally, qualitative examples further illustrate how LLMs reason over mmWave point cloud inputs.

The proposed mmWave-QA benchmark evaluates commercial LLMs \citep{openai2024gpt4o, deepmind2025gemini25} reasoning on radar-based human activity data collected under heterogeneous devices, environments, and scene conditions.
Quantitative results under different prompting strategies are presented first, along with analyses comparing mmWave- and RGB-based QA and examining the effect of frame count on performance. 
Finally, qualitative examples further illustrate how LLMs reason over mmWave point cloud inputs.
Detailed evaluation settings are provided in the supplementary material.

%\subsection{Settings}
%The evaluation involves four commercial models, i.e., GPT-4o \citep{openai2024gpt4o}, GPT-5, Gemini 2.5-flash, and Gemini 2.5-pro \citep{deepmind2025gemini25}.
%For fair comparison, the number of frames is fixed to 8 per sample for both mmWave and RGB inputs.
%Accuracy is computed by directly comparing the model’s output with the ground-truth answer, without utilizing any external models such as GPT.

\subsection{Quantitative Results}
\paragraph{Q: Can LLMs understand radar point clouds without task-specific training?}  
%As shown in Table~\ref{tab:action_category_comparison}, the performance progressively improves from zero-shot to few-shot, CoT, and finally few-shot CoT prompting, consistently across all models.
To explore this question, we evaluate various prompting strategies, where the results in Table~\ref{tab:action_category_comparison} show consistent performance gains-from zero-shot to few-shot, CoT, and finally few-shot CoT prompting-across all models.
This behavior highlights the complementary benefits of in-context learning and rationale-based reasoning, demonstrating their potential to enhance understanding of radar-derived motion representations.
GPT-5 achieved the best overall performance in the few-shot CoT setting, improving overall accuracy by 1.39 and 1.86 percentage points over GPT-4o and Gemini 2.5-Pro, respectively.
Among action categories, arm motions yield the best results, followed by full-body movements, while leg and torso actions remain challenging due to subtler motion cues and frequent occlusions in radar reflections. 
Across mmWave hardware, Phoenix shows the highest accuracy owing to its high-quality point clouds enabled by an advanced radar antenna configuration, whereas IWR6843 and IWR1443 show performance degradation caused by sensor noise and environmental variability.
These results provide empirical evidence that LLMs can reason robustly over heterogeneous radar datasets, suggesting their strong generalization capability even under diverse sensing conditions.

\begin{table*}[t]
\centering
\small
\renewcommand{\arraystretch}{0.5}
\setlength{\tabcolsep}{3pt}
\begin{tabular}{>{\raggedright\arraybackslash}p{2.2cm}>{\raggedright\arraybackslash}p{2cm}>{\centering\arraybackslash}p{0.7cm}>{\centering\arraybackslash}p{0.7cm}>{\centering\arraybackslash}p{0.7cm}>{\centering\arraybackslash}p{0.7cm}>{\centering\arraybackslash}p{0.8cm}>{\centering\arraybackslash}p{0.8cm}>{\centering\arraybackslash}p{0.8cm}>{\centering\arraybackslash}p{1cm}>{\centering\arraybackslash}p{1.1cm}>{\centering\arraybackslash}p{1cm}>{\centering\arraybackslash}p{0.7cm}>{\centering\arraybackslash}p{0.7cm}}
\toprule
\multirow{2}{*}{\textbf{Model}} & \multirow{2}{*}{\textbf{Prompt Type}} & \multicolumn{4}{c}{\textbf{ActNum}} & \multicolumn{3}{c}{\textbf{ActOrder}} & \multicolumn{3}{c}{\textbf{LimbFocus}} & \multicolumn{2}{c}{\textbf{TrajCheck}}\\ 
\cmidrule(lr){3-6} \cmidrule(lr){7-9} \cmidrule(lr){10-12} \cmidrule(lr){13-14}
 &  & 1 & 2 & 3 & 4 & Prev & Present & Next & Single & Arm/Leg & Multiple & Static & Move\\ 
\midrule
\multirow{4}{*}{\textbf{Gemini2.5-flash}} 
 & Zero-shot    & 9.82           & 47.50          & 5.00           & 7.50  & 24.80 & 26.88 & 27.20 & 21.67 & 24.80 & 28.00 & 3.20           & 87.20\\
 & Few-shot     & 38.39          & 52.50          & 36.67          & 17.50 & 8.00  & 30.00 & 32.00 & 30.00 & 31.20 & 38.00 & 27.20          & 43.20\\
 & CoT          & 11.60          & 50.00          & 30.00          & 15.00 & 26.40 & 28.75 & 28.80 & 23.33 & 25.60 & 32.00 & 7.20           & \textit{88.00} \\
 & Few-shot-CoT & \textit{41.07} & 55.00          & \textbf{45.00} & \textbf{22.50} & \textit{30.40} & \textbf{30.63} & \textit{35.20} & \textit{35.00} & \textit{33.60} & \textbf{42.00} & \textit{62.40} & 70.40 \\
 \midrule
 \multirow{4}{*}{\textbf{GPT-4o}} 
 & Zero-shot    & 11.61          & 52.50          & 5.00           & 5.00 & 24.80 & 26.88 & 27.20 & 21.67 & 24.80 & 28.00 & 4.00           & \textbf{92.00} \\
 & Few-shot     & 40.18          & \textit{56.25} & 35.00          & 17.50 & 28.80 & 28.75 & 31.20 & 26.67 & 28.80 & 38.00 & 32.00          & 44.00\\
 & CoT          & 30.36          & 36.25          & 25.00          & 12.50 & 26.40 & 27.50 & 29.60 & 25.00 & 26.40 & 34.00 & 10.40          & \textit{88.00}\\
 & Few-shot-CoT & \textbf{42.86} & \textbf{58.75} & \textit{43.33} & \textbf{22.50} & \textbf{31.20} & \textbf{30.63} & \textbf{36.00} & \textbf{38.33} & \textbf{35.20} & \textbf{42.00} & \textbf{68.00} & 72.80\\
\bottomrule
\end{tabular}
\caption{Evaluation of various LLMs across different prompt settings, including ActNum, ActOrder, LimbFocus, and TrajCheck QA. 
    All values denote accuracy (\%), with the best results in \textbf{bold} and the second-best in \textit{italic}.}
\label{tab:comparison_no_overall}
\end{table*}

\begin{figure*}[t] 
  \centering
  \includegraphics[width=\linewidth]{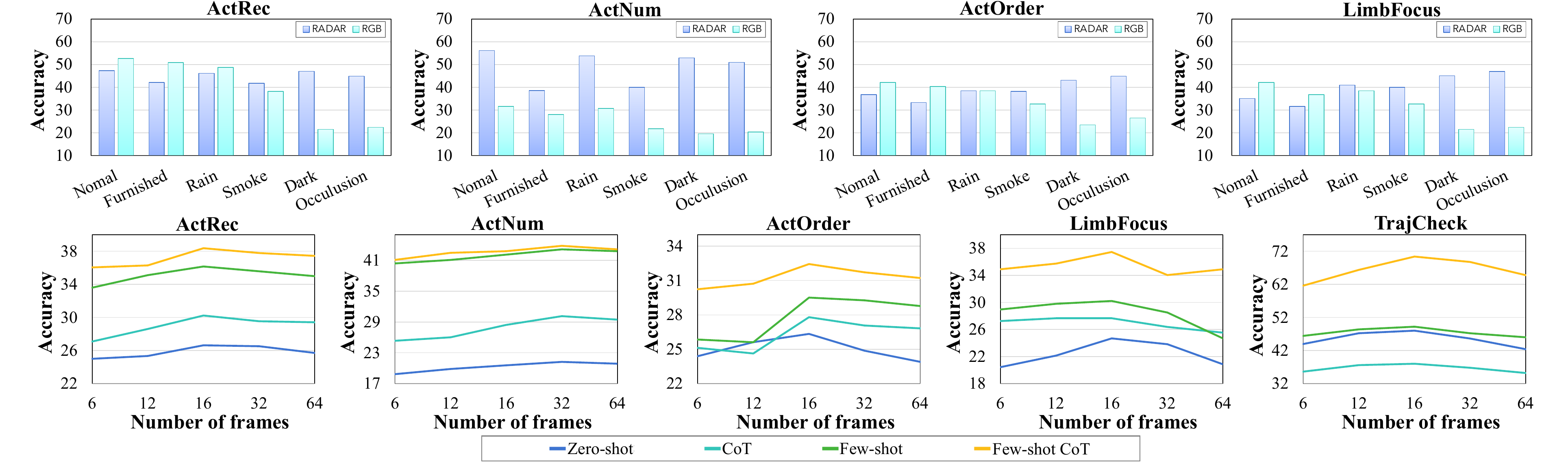} 
  \caption {\textbf{Top:} Comparison of mmWave- and RGB-based QA performance across six scenarios: normal, furnished, rain, smoke, dark, and occlusion. 
    \textbf{Bottom:} Performance variation with respect to the number of frames used in each QA task.}
  \label{fig:graph_fig}
\end{figure*}

\vspace{-10pt}
\paragraph{Q: How effectively do LLMs perform multi-aspect reasoning across human-motion QA tasks?}  
We extend our evaluation to multi-aspect reasoning QA tasks, as shown in Table~\ref{tab:comparison_no_overall}. 
For TrajCheck in the zero-shot setting, both GPT-4o and Gemini 2.5-flash predominantly classify radar samples as ``move'', achieving 92.0\% and 87.2\% accuracy for moving cases but only 4.0\% and 3.2\% for static ones.
When few-shot examples are provided, the static accuracy increases to 32.0\% and 27.2\%, respectively, suggesting that contextual information helps both models better distinguish between moving and static cases.
For ActNum, even under the few-shot CoT setting, models achieve stable accuracy for 2–3 actions at 43–58.75\%, but sharply decline to 5-22.5\% as the action count increases, indicating difficulty in sustaining consistent reasoning over extended temporal sequences in point clouds.
In ActOrder, predictions for the present are most accurate, whereas reasoning about previous and next actions remains challenging, likely because the excessive number of input tokens from dense point clouds hinders the model’s ability to capture clear temporal relationships.
Finally, LimbFocus demonstrates strong recognition for single or limb-specific motions but lower accuracy in multi-limb scenarios, 
likely due to multipath interference and ghost reflections that distort body-part separability in complex radar scenes.
These findings promote future studies on understanding the structural composition of radar point clouds and selecting temporally salient frames to facilitate more effective temporal reasoning.

\begin{figure*}[t] 
  \centering
  \includegraphics[width=\linewidth]{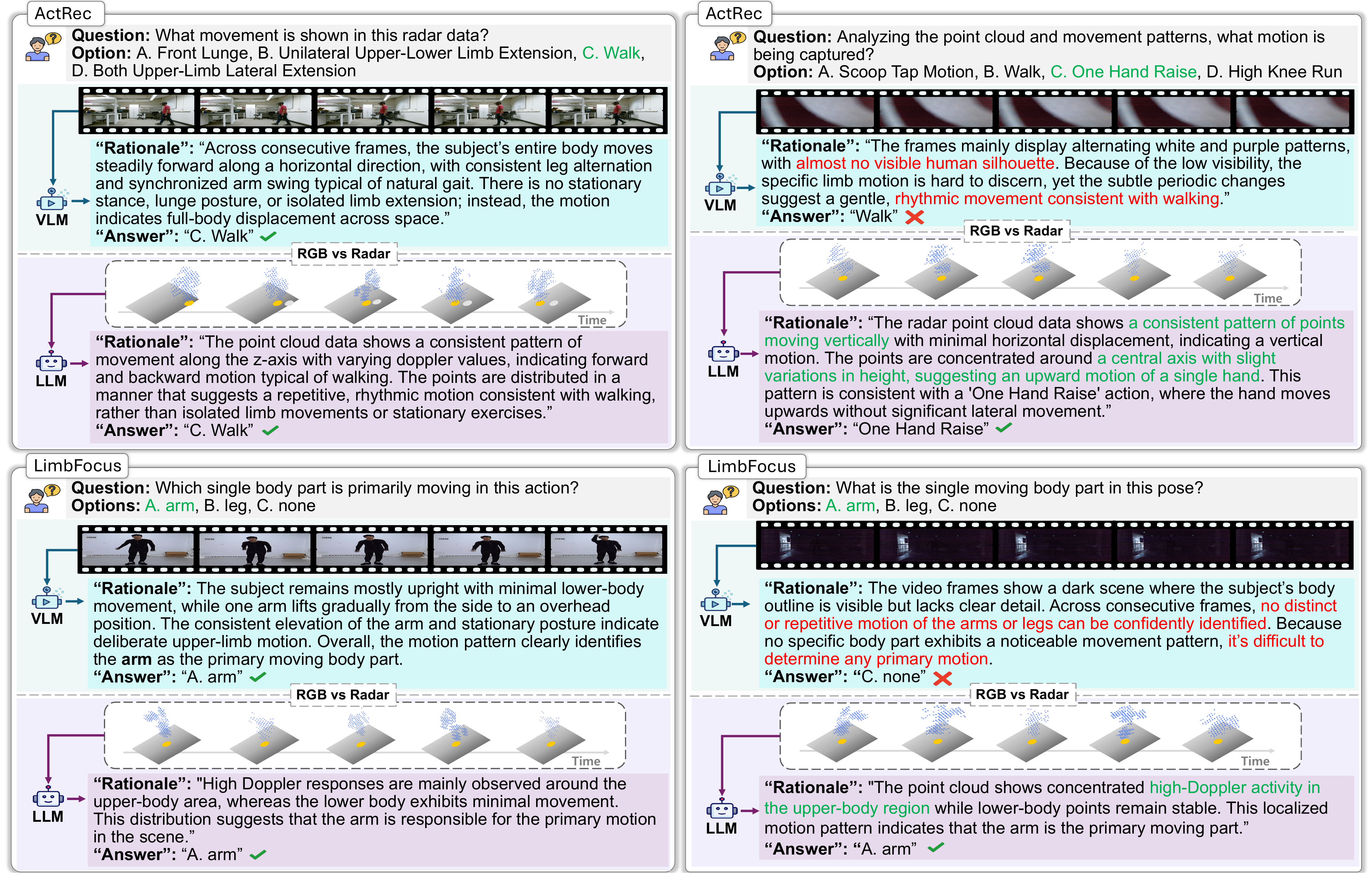} 
  \caption { \textbf{Qualitative analysis of mmWave-QA.}
    Each example from the ActRec and LimbFocus tasks visualizes temporal changes in video and radar point clouds along with corresponding rationales generated by the VLM and the LLM.
    When both modalities provide clear visual evidence, their reasoning remains consistent; however, under degraded conditions (\eg, occlusion or darkness), the VLM often hallucinates or misinterpret corrupted frames, whereas radar-based reasoning remains stable and physically grounded.
    }
  \label{fig:qualitative}
\end{figure*}

\subsection{Ablation Studies}
\paragraph{Comparison with RGB Modality.}
We compare the performance of mmWave-based QA with its RGB counterpart across six scenarios, as shown in the upper part of \cref{fig:graph_fig}.
For the RGB modality, we evaluate using GPT-4o \citep{openai2024gpt4o}, which accepts RGB inputs.
In normal and furnished environments, RGB outperforms radar, as vision models have been trained on far larger and more diverse datasets, where richer visual cues facilitate easier reasoning for VLMs, whereas multipath ghost reflections from furniture degrade the quality of radar point clouds.
However, in rain and smoke conditions, both modalities achieve comparable performance, demonstrating radar’s robustness against visibility degradation. 
Notably, in dark and occlusion scenarios, the radar modality surpasses RGB, highlighting its unique advantage in perception tasks where visual information is heavily corrupted or unavailable.
In contrast to other tasks, ActNum shows consistently better performance with radar, as motion counting relies on temporal Doppler cues that are hard for RGB models to capture with limited frames.

%In contrast to other tasks, ActNum demonstrates superior performance with radar, suggesting that motion-counting tasks benefit from the temporal continuity and velocity information embedded in radar signals.
%This suggests that RGB models struggle to distinguish subtle motion differences with limited frames, whereas radar provides more effective reasoning through Doppler velocity and coordinate variation cues.

\vspace{-6pt}
\paragraph{Frame-wise Performance Analysis.}
As shown in the lower part of \cref{fig:graph_fig}, the performance varies with the number of frames used in each QA task, where we gradually increase the frame count from 6 to 64 during evaluation.
The input frame length plays a critical role in interpreting human activity from point cloud sequences, as it determines both temporal coverage and information density.
Across most tasks, performance peaks around 16 frames, as too few frames limit temporal context, while excessive frames introduce redundant or noisy information that hinders reasoning.
This finding suggests that radar-based motion reasoning requires an optimal temporal balance--segments should be long enough to capture key motion patterns but short enough to avoid redundant or noisy points.

\subsection{Qualitative Analysis}
To qualitatively examine reasoning behavior, \cref{fig:qualitative} illustrates the rationales and responses produced by the VLM and LLM given video and point cloud inputs.
When RGB provides clear visual evidence, its interpretations are largely consistent with radar, producing similar answers and reasoning patterns.
However, under visually degraded conditions such as darkness or occlusion, the RGB-based model often exhibits hallucinated reasoning or over-explains corrupted regions, leading to incorrect predictions (\eg, “almost no visible human”).
In contrast, the radar modality facilitates Doppler-based motion reasoning, proving effective when visual information is degraded or unavailable.
In ActRec, the LLM correctly infers walking motion by tracking coordinate displacement and Doppler variation across frames (\eg, “high-Doppler activity in the upper-body region”).
That is, LLM reasoning relies on coordinate and Doppler cues from radar inputs, while VLM depends on visual features, as detailed in the supplementary material.
%LLM reasoning leverages coordinate and Doppler cues from radar inputs, while VLM relies on visual features.
%Additional qualitative analyses for all QA tasks are provided in the Supplementary Material.

%To qualitatively examine model reasoning, \cref{fig:qualitative} illustrates the rationales and responses generated by the VLM and LLM models given video and point cloud inputs.
%When RGB provides clear visual evidence, its interpretations are largely consistent with radar, producing similar answers and reasoning patterns.
%However, under visually degraded conditions such as darkness or occlusion, the RGB-based model often exhibits hallucinated reasoning or over-explains corrupted regions, leading to incorrect predictions.
%In contrast, the radar modality maintains stable reasoning grounded in motion and Doppler cues, demonstrating its robustness to visual corruption.
%LLM reasoning leverages coordinate and Doppler cues from radar inputs, while VLM relies on visual features.
%Additional qualitative analyses for all QA tasks are provided in the Supplementary Material.

\vspace{-1pt}
\section{Conclusion}
\label{sec:conclusion}

In this paper, we present the first mmWave-QA benchmark designed to assess a diverse range of human actions through natural language QA.  
Our benchmark integrates multiple mmWave sensing datasets to capture heterogeneity across devices, environments, and scene conditions through hardware-aware preprocessing and global taxonomy.
Through extensive experiments, we demonstrate that LLMs can effectively interpret mmWave point cloud inputs and perform effective QA tasks.
Notably, with the same underlying model, LLMs conditioned on mmWave point cloud inputs demonstrate greater robustness than those conditioned on RGB inputs, especially under visually degraded conditions such as low light or occlusion. 
This finding underscores the potential of leveraging LLMs for mmWave understanding to achieve reliable reasoning in vision-limited environments.
While our study demonstrates the promising capabilities of LLMs for radar-based understanding using off-the-shelf models such as GPT and Gemini, exploration with diverse open-source LLMs remains limited.
Future work will focus on improving open-source models through fine-tuning and exploring broader integration with reasoning-oriented frameworks such as retrieval-augmented generation systems and intelligent agents.
We hope our work paves the way toward a broader understanding of multimodal reasoning that bridges non-visual sensing with language-based intelligence.

{
    \small
    % Use a TeX Live-provided style so BibTeX also works in isolated build steps.
    \bibliographystyle{abbrvnat}
    \bibliography{main}
}

\end{document}